\documentclass[11pt,a4paper]{article}
\usepackage{authblk}

\usepackage[hyperref]{emnlp-ijcnlp-2019}
\usepackage{times}
\usepackage{latexsym}
\usepackage{graphicx}
\usepackage{color}
\usepackage{url}
\usepackage{caption,subcaption}
\usepackage{amsmath}
\usepackage{amssymb}

\aclfinalcopy 

\aclfinalcopy 

\title{ParaQG: A System for Generating Questions and Answers from Paragraphs}
\author[1,3,4]{Vishwajeet Kumar}
\author[2]{Sivaanandh Muneeswaran}
\author[3]{Ganesh Ramakrishnan}
\author[4]{Yuan-Fang Li}
\affil[1]{IITB-Monash Research Academy, Mumbai, India}
\affil[2]{Mepco Schlenk Engineering College, Tamilnadu, India}
\affil[3]{IIT Bombay, Mumbai, India}
\affil[4]{Monash University, Melbourne, Australia}

\date{}
\begin{document}
\maketitle
\begin{abstract}
Generating syntactically and semantically valid and relevant questions from paragraphs is useful with many applications. Manual generation is a labour-intensive task, as it requires the reading, parsing and understanding of long passages of text. A number of question generation models based on sequence-to-sequence techniques have recently been proposed. Most of them generate questions from sentences only, and none of them is publicly available as an easy-to-use service. In this paper, we demonstrate ParaQG, a Web-based system for generating questions from sentences and paragraphs. ParaQG incorporates a number of novel functionalities to make the question generation process user-friendly. It provides an interactive interface for a user to select answers with visual insights on generation of questions. It also employs various faceted views to group similar questions as well as filtering techniques to eliminate unanswerable questions. 
\end{abstract}

\section{Introduction}
Asking relevant and intelligent questions has always been an integral part of human learning, as it can help assess user understanding of a piece of text (a comprehension, an article, etc.). 
However, forming questions manually is an arduous task. 
Automated question generation (QG) systems can help alleviate this problem by learning to generate questions on a large scale efficiently. 
A QG system has many applications in a wide variety of areas such as FAQ generation, intelligent tutoring systems, automating reading comprehension, and virtual assistants/chatbots. 
For a QG system, the task is to generate syntactically coherent, semantically correct and natural questions from text. 
Additionally, it is highly desirable that the questions are relevant to the text and are pivoted on answers present in the text.
\begin{figure}[!htb]
\centering
\includegraphics[width=\linewidth,trim={2.4cm 4.2cm 2.4cm 8mm}, clip]{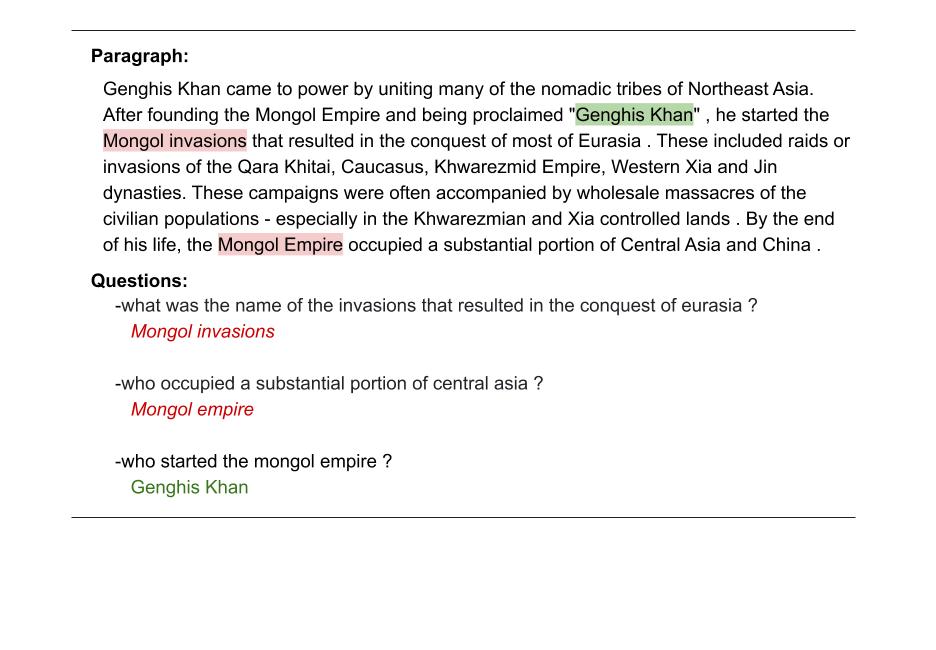}
\caption{Example: Questions generated from the same paragraph across choices of pivotal answer(s).}
\label{ex1}
\end{figure}
Distinct from other natural language generation tasks such as summarisation and paraphrasing, answers play an important role in question generation. 
Different questions can be formed from the same passage based on the choice of the \emph{pivotal answer}. The \emph{pivotal answer} is the span of text from the input passage around which a question is generated. The \emph{pivotal answer} can be either selected manually by the user,  automatically by the system or by a combination of the two. For example in Figure \ref{ex1} it can be seen that based on different answers selected (highlighted in different colours), our system generates different questions.
\par Neural network-based sequence-to-sequence (Seq2Seq) models represent the state-of-the-art in question generation. 
Most of these models~\cite{xinya,kumar18,song2018leveraging,Kumar2019PHBC,Kumar2019CrossLingualTF} take single sentence as input, thus limiting their usefulness in real-world settings. Some recent techniques tackle the problem of question generation from paragraphs~\cite{zhao2018paragraph}. 
However, none of the above works is publicly available as an online service. 

In this work we present ParaQG, an interactive Web-based question generation system to generate correct, meaningful and relevant questions from sentences, and paragraphs.
Given a passage of text as input, users can manually select a set of answer spans to ask questions about ({\em i.e.} choose answers) from an automatically curated set of noun phrases and named entities. 
Questions are then generated by a combination of a (novel) sequence-to-sequence model with dynamic dictionaries, the copy mechanism~\cite{gu2016incorporating} and the global sparse-max attention~\cite{martins2016softmax}. 

ParaQG incorporates the following main features. 
\begin{enumerate}
\item An interactive, user-configurable Web application to automatically generate questions from a sentence, or a paragraph based on user selected answers, with visual insights on the generated questions. 
\item A technique to create faceted views of the generated questions having overlapping or similar answers. 
Given an input passage, the same answer may appear multiple times in different spans, from which similar questions can be generated. 
ParaQG detects and presents  the generated questions based on a grouped/faceted view of similar answer spans, thus allowing easy selection by  users. 
\item A novel question filtering technique based on BERT~\cite{devlin2018bert} to eliminate unanswerable questions from the text.
 \end{enumerate}
To the best of our knowledge we are the first to propose and develop an interactive system that generates questions based on the answers selected by users. 
The rest of the paper is organized as follows. We discuss related work in Section \ref{relwork}. In Section \ref{systemarch}, We describe the architecture of ParaQG. This is followed by details of the demonstration in Section \ref{demodetails} and the implementation in Section \ref{sec:impl}. Conclusion is discussed in Section \ref{conc}.

\section{Related Work}
\label{relwork}
Automatically generating questions and answers from text is a challenging task. This task can be traced back to 1976 when Wolfe~\shortcite{wolfe1976automatic} presented their system AUTOQUEST, which examined the generation of Wh-questions from single sentences. This was followed by several pattern matching \cite{hirschman1999deep} and linear regression \cite{ng2000machine} based models. These approaches are heavily dependent on either rules or question templates, and require deep linguistic knowledge, yet are not exhaustive enough. Recent successes in neural machine translation \cite{sutskever2014sequence,cho2014properties} have helped address these issues by letting deep neural nets learn the implicit rules from data. This approach has inspired application of sequence-to-sequence learning to automated question generation. Serban et al.~\shortcite{serban2016generating} proposed an attention-based \cite{bahdanau2014neural,LuongPM15} approach to question generation from a pre-defined template of knowledge base triples (subject, relation, object). We proposed multi-hop question generation \cite{Kumar2019Difficulty} from knowledge graphs using transformers~\cite{vaswani2017attention}. Du et al.~\shortcite{xinya} proposed an attention-based sequence learning approach to question generation.

Most existing work focuses on  generating questions from text without concerning itself with answ er generation. In our previous work~\cite{kumar18}, we presented a pointer network-based model that predicts candidate answers and generates a question by providing a pivotal answer as an input to the decoder. Our model for question generation combines a rich set of linguistic features, pointer network-based answer selection, and an improved decoder, and is able to generate questions that are relatively more relevant to the given sentence than the questions generated without the answer signal. 

Overall, the broad finding has been that it is important to  either {\em be provided with} or {\em learn to choose} pivotal answer spans to ask questions about from an input passage. Founded on this observation, our system facilitates users with an option to either choose answer spans from the pre-populated set of named entities and noun phrases or manually select custom answer spans interactively.
Our system, ParaQG, presented in this paper uses a novel four-stage procedure: (1) text review, (2) pivotal answer selection (3) automatic question generation pertaining to the selected answer, and (4) filtering and grouping questions based on confidence scores and different facets of the selected answer.

\section{System Architecture}
\label{systemarch}
ParaQG generates questions from sentences and paragraphs following a four-stage interactive procedure: (a) paragraph review, (b) answer selection, (c) question generation with associated confidence score, and (d) question filtering and grouping based on answer facets. 
Given a paragraph, ParaQG first reviews the content automatically and then flags any unprocessable characters (e.g.\ Unicode characters) and URLs, which the user are prompted to edit or remove (Section~\ref{qwss}). Next, the user is provided with an option to select an answer from the list of candidate answers identified by the system. Alternatively, the user can select custom answer spans from the passage to ask question about (Section \ref{sec:as}). In the third step, the selected pivotal spans are encoded into the paragraph and fed to the question generation module. The question generation module is a sequence-to-sequence model with dynamic dictionaries, reusable copy attention and global sparse-max attention. This module attempts to automatically generate the most relevant as well as syntactically and semantically correct questions around the selected pivotal answers (Section \ref{sec:qg}). In the last step the unanswerable questions are filtered out using a BERT-based question filtering module (Section \ref{sec:filtering}). The questions that remain are presented by grouping their associated answers. Each group of answers (which we also refer to as an answer-based facet) corresponds to some unique stem form of those answer words.

\subsection{Paragraph Review}
\label{qwss}
\begin{figure*}[htb]
\centering 
\begin{subfigure}{0.4\textwidth}
    \centering
    \includegraphics[width=\linewidth]{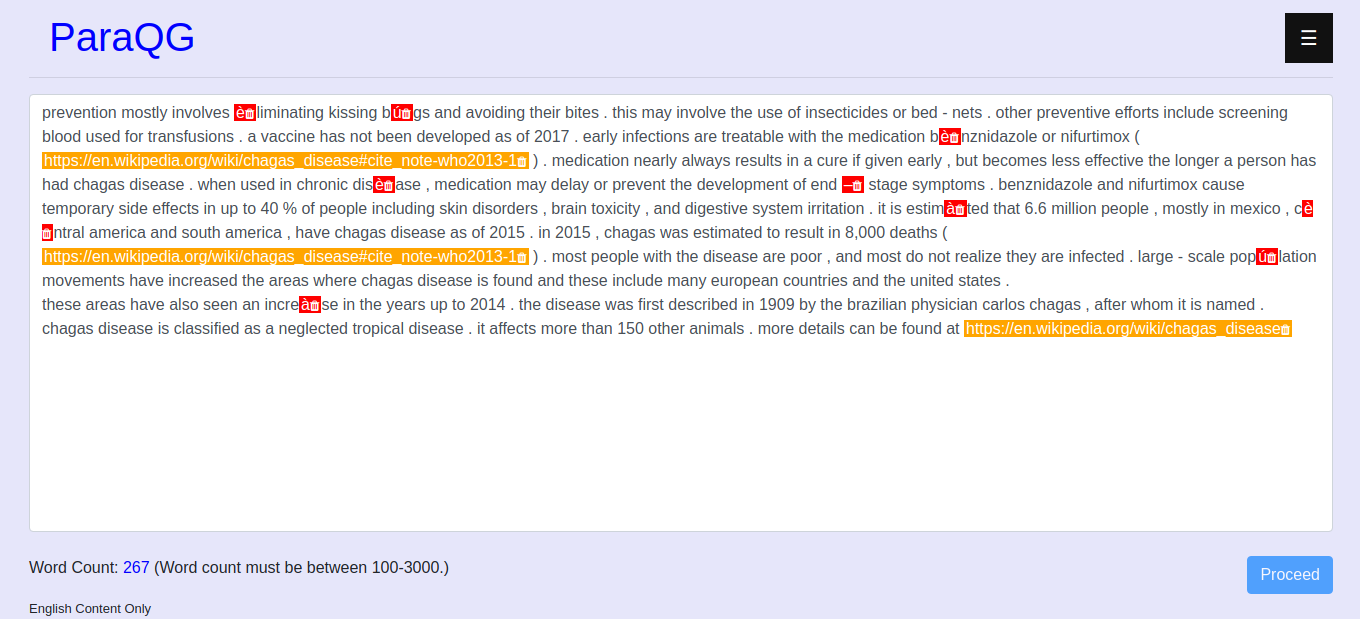}
    \caption{Reviewing paragraph content.}
    \label{fig:review}
\end{subfigure}\hfil 
\begin{subfigure}{0.4\textwidth}
  \includegraphics[width=\linewidth]{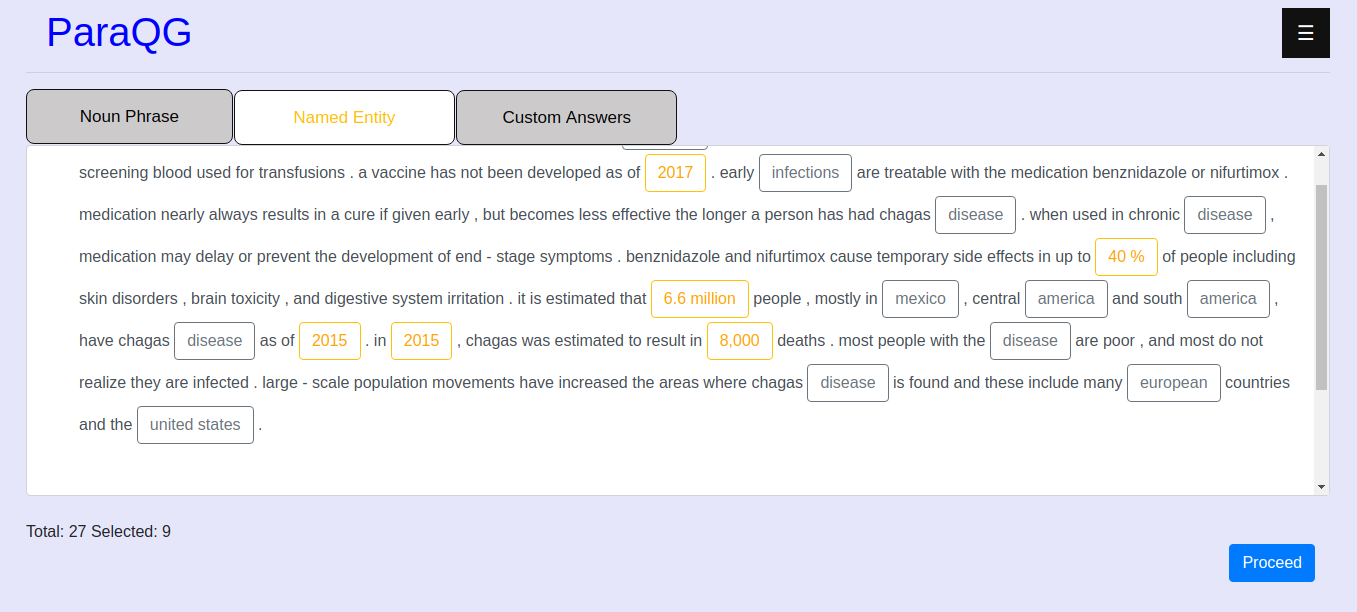}
  \caption{Selecting pivotal answers from named entities.}
  \label{fig:ne-as}
\end{subfigure}\hfil 
\begin{subfigure}{0.4\textwidth}
  \includegraphics[width=\linewidth]{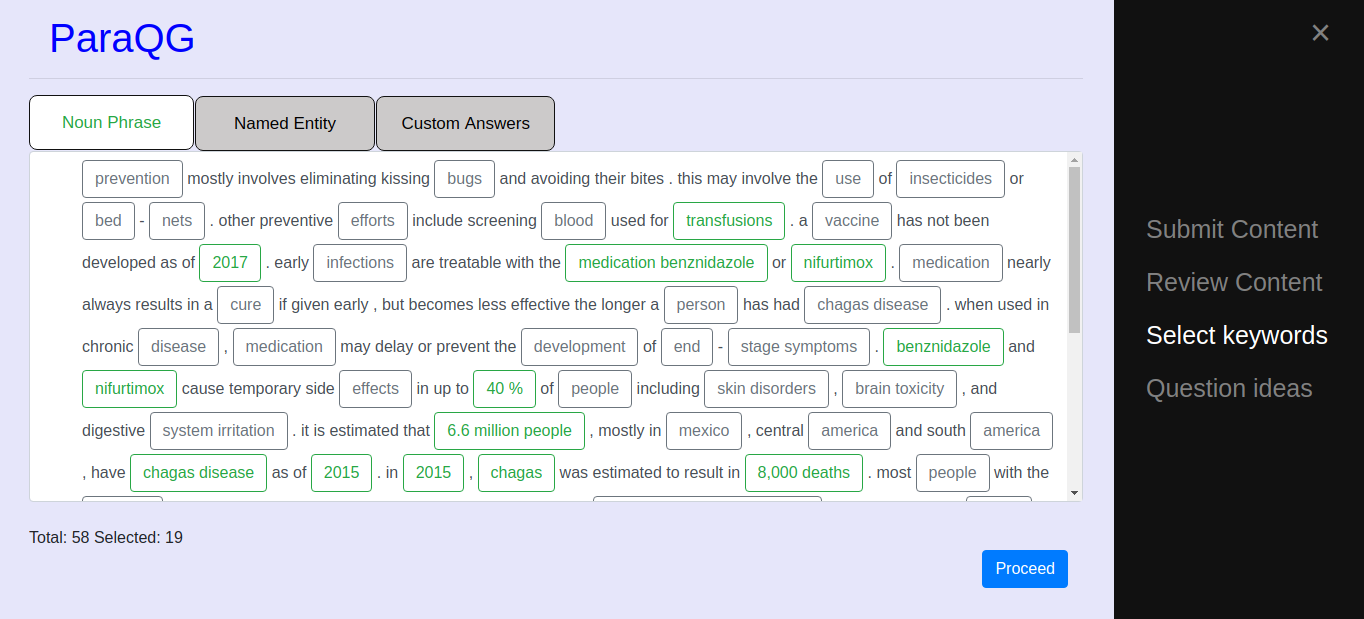}
  \caption{Selecting pivotal answers from noun phrases.}
  \label{fig:np-as}
\end{subfigure}\hfil 
\begin{subfigure}{0.4\textwidth}
    \centering
    \includegraphics[width=\linewidth]{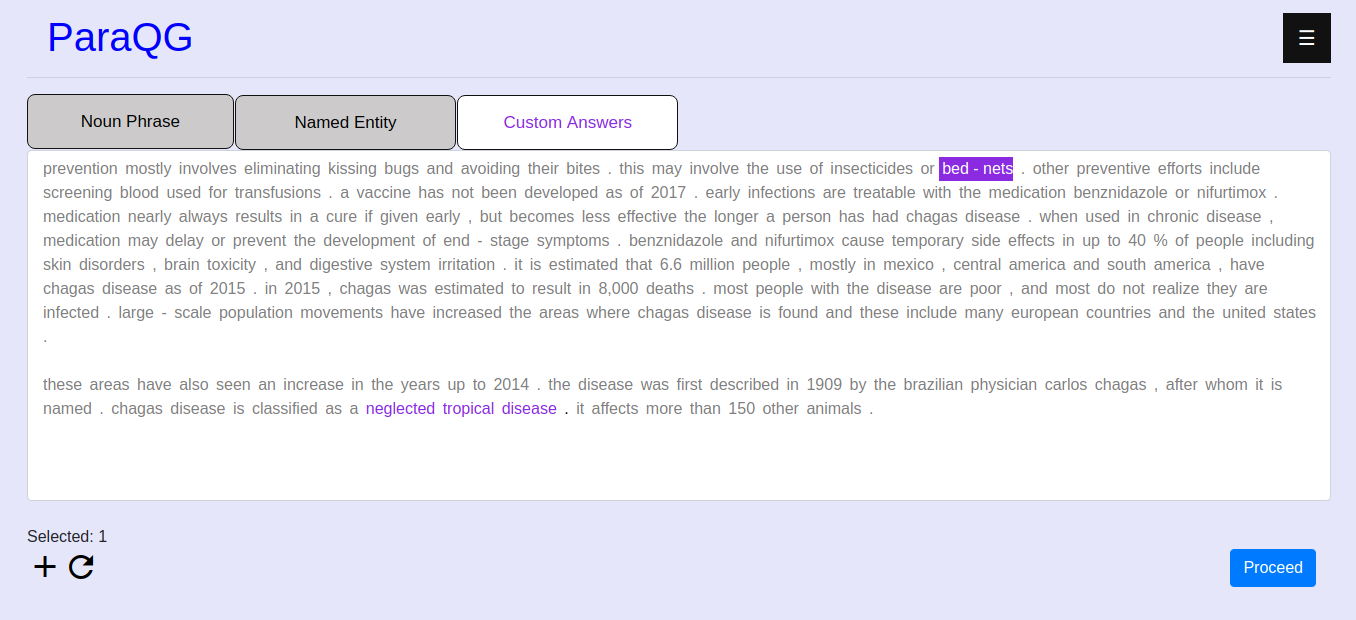}
    \caption{Interactive pivotal answer selection.}
    \label{fig:ans_sel}
\end{subfigure}\hfil 
\caption{Main steps of ParaQG.} 
\end{figure*}

Since every sentence/word in a paragraph may not be question-worthy, it is important to filter out those that are not. Given a paragraph text, the system automatically reviews its contents to check if the paragraph contains any non-ASCII characters, URLs {\em etc.}, and flags them for users to edit, as illustrated in Figure~\ref{fig:review}.

\subsection{Answer Selection}
\label{sec:as}
ParaQG allows users to select any named entity or noun phrase present in the paragraph as a \emph{pivotal answer}. As mentioned earlier, a user is presented with a list of all the named entities and noun phrases as extracted using the Stanford CoreNLP tagger to  choose pivotal answers from. Alternatively, users can manually select a set of spans from the passage as pivotal answers, as shown in Figure~\ref{fig:ans_sel}. The selected answer is encoded in the source sentence using the BIO (Begin, Inside, Outside) notation. 

\subsection{Question Generation}
\label{sec:qg}
Similar to our previous work \cite{kumar18}, we encode the pivotal answer spans in the passage using BIO notation, and train a sequence-to-sequence model augmented with dynamic dictionary, copy mechanism and global sparse-max attention. Our question generation module consists of a paragraph encoder and a question decoder. The encoder represents the paragraph input as a single fixed-length continuous vector. This vector representation of the paragraph is passed to the decoder with reusable copy mechanism and sparse-max attention to generate questions. 

\subsection{BERT-based Question Filtering}
\label{sec:filtering}
We use the $BERT_{base}$ \cite{devlin2018bert} model to filter out unanswerable questions generated by our model. we fine-tune BERT on SQuAD 2.0\cite{rajpurkar2018know}. SQuAD 2.0 extends SQuAD with over 50000 unanswerable questions. The unanswerable questions are flagged with the attribute \textit{is\_impossible}.  

We represent input question (question generated by our QG model) and the passage in a single packed sequence of tokens, while using a special token [SEP] to separate the question from the passage. Similar to \cite{devlin2018bert} we use a special classification token [CLS] at the start of every sequence. Let us denote the final hidden representation of the [CLS] token by $C$ and the final hidden representation for the $i^{th}$ input token by $T_i$. For each unanswerable question, we represent the start and end answer index using a [CLS] token as it does not have any answer start and end index. Similar to \cite{devlin2018bert}, we compare the score of no-answer span with the score of best non-null answer span to predict the answerability of a question.
Score of no-answer span is calculated as: $s_{null} = S.C + E.C$
where $S \in \mathbb{R}^H $ is the vector representation of answer start index and $E \in \mathbb{R}^H$ is the vector representation of answer end index. 
The score of a non-null answer span is defined as $s_{i,j} = \max_{j>=i} \{S.T_i + E.T_j\}$
If the score of $s_{null} - s_{i,j}> V$, where $V$ is a threshold calculated using a validation set, then the question is not answerable using the paragraph.

\subsection{Grouped/Faceted Views of Questions}
We group together all answers and their corresponding question(s) that have the same stemmed form. For example, two potential answer spans `switching' and `switches' would have the same stemmed form `switch'. Thus, the spans `switching' and `switches', and their associated question(s) would be grouped together under the same stemmed form `switch'. 
Summarily, each such question group yields a {\em faceted view} of the question set. Within each group, the questions are sorted in decreasing order of their probabilities. 
We calculate the \emph{intra-question probability (confidence score)} by normalizing the beam score $x$ as: $\frac{e^x}{1+ e^x}$.
The final \emph{inter-question probability} of a question-answer pair is calculated from the question with maximum intra-question probability $p$ as: $\frac{p-min(\textbf{P})}{max(\textbf{P})-min(\textbf{P})}$, where \textbf{P} is the set of maximum probability scores across answers.

\section{Demonstration Details}
\label{demodetails}
ParaQG is available as an interactive and fully-featured Web application. A video of the ParaQG system is available at \url{https://youtu.be/BLChd18kz1c}. The ParaQG system is accessible at \url{https://www.cse.iitb.ac.in/~vishwajeet/paraqg.html}.
Important features of the Web application are discussed below.
\paragraph{Input and content review:} A user can copy any paragraph and paste it in the text area (Fig.~\ref{fig:review}), and subsequently will be asked to review and remove/edit unprocessable contents (Fig.~\ref{fig:review}). 

\paragraph{Interactive pivotal answer selection:} ParaQG provides an interactive user interface for users to select pivotal answers. A user has an option to select a pivotal answers either from a set of noun phrases or from a set of named entities present in the paragraph. To choose a pivotal answer from a set of named entities, the user can click on the \textit{Named Entities} tab (Figure \ref{fig:ne-as}). Similarly, to select a noun phrase present in the paragraph as the pivotal answer, the user can click on the \textit{Noun Phrases} tab (Figure \ref{fig:np-as}). Once a user clicks on either of the tabs he/she will be presented with pre-highlighted noun phrases/named entities as pivotal answers. The user can subsequently deselect a pivotal answer by clicking on it. 

\paragraph{Custom pivotal answer selection:} Alternatively, the user can click on the \textit{Custom Answers} tab (Figure \ref{fig:ans_sel}) and manually select the most important spans in the paragraph as the pivotal answers. The users can also select overlapping spans.

\begin{figure}[t]
    \centering
    \includegraphics[width=\linewidth]{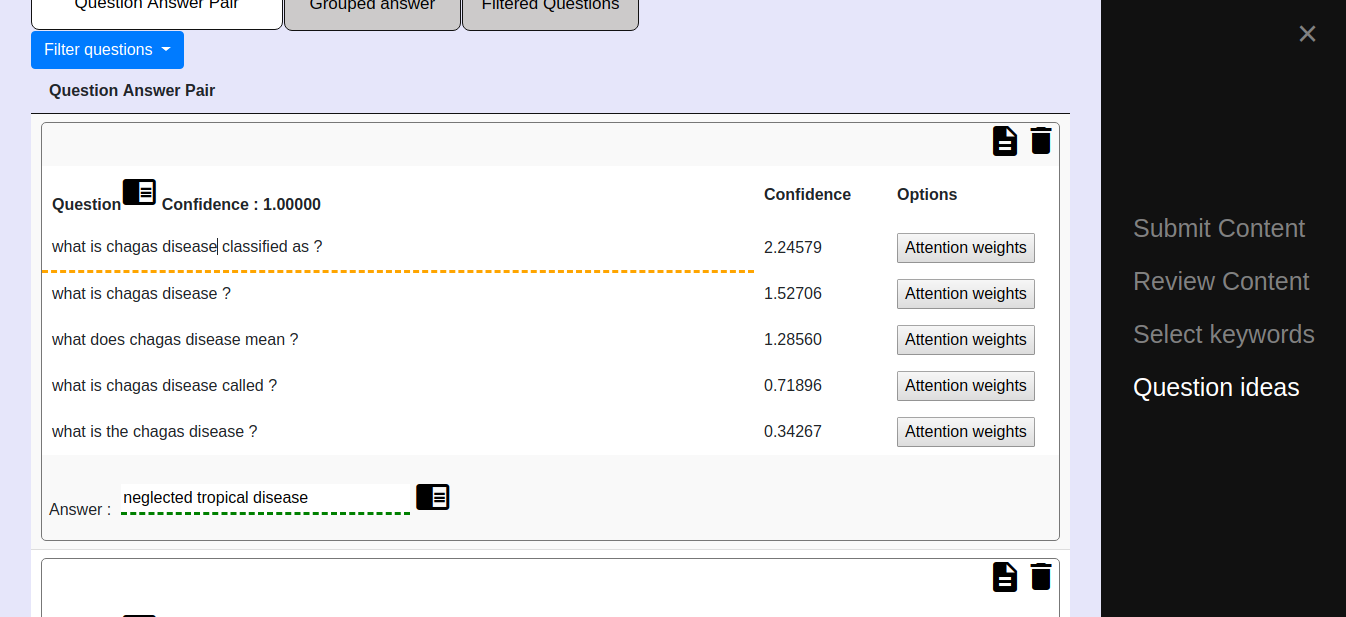}
    \caption{Editing question and answers.}
    \label{fig:edit_question}
\end{figure}

\paragraph{Automatic question and answer generation:} Finally, the user is presented with the question generated as well as the answer to that question along with confidence score. For example for the paragraph input by the user in Figure \ref{fig:ne-as}, the questions as well as the answers are generated and shown to the the user (Figure \ref{fig:edit_question}) along with their confidence score.

\begin{figure}[t]
    \centering
    \includegraphics[width=\linewidth]{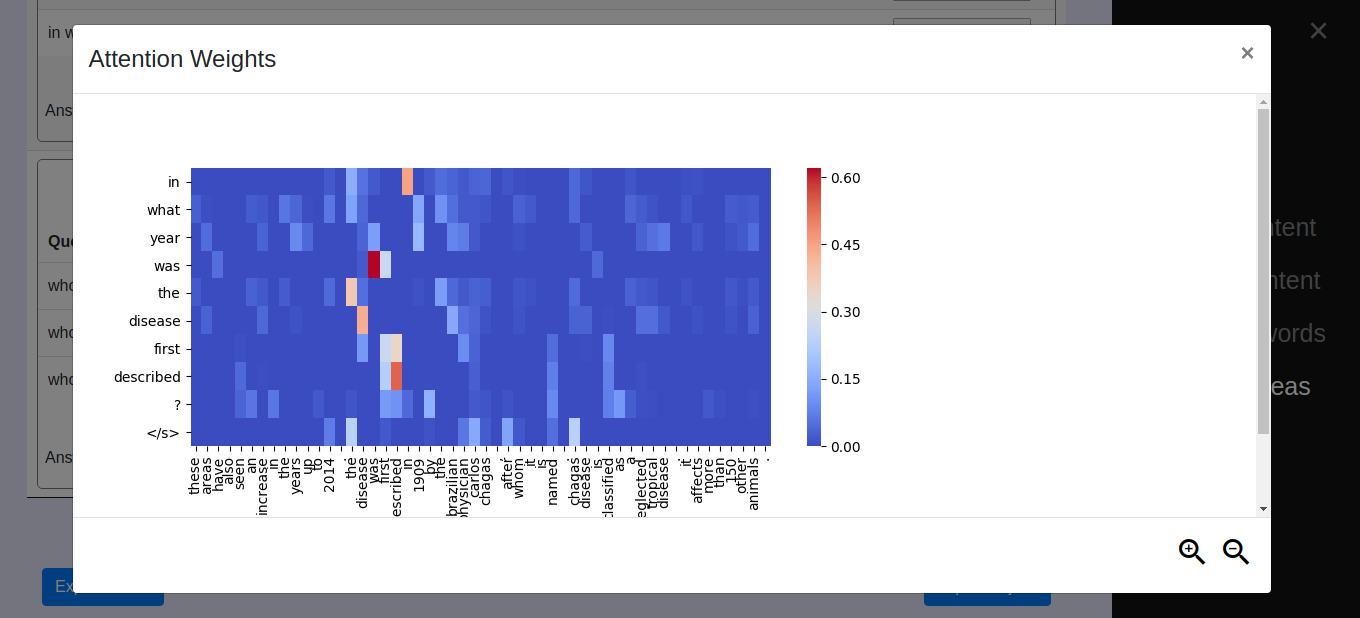}
    \caption{Attention weight visualization.}
    \label{fig:attn_vis}
\end{figure}

\paragraph{Visualization of decoder attention weights using heat maps:} ParaQG also presents to the user heat maps of the decoder attention weights between words in the paragraph and words in the question generated. A user can click on the \textit{attention weights} button next to each question (Figure \ref{fig:edit_question}) to generate the attention weights heat map between words in the paragraph and words in the generated question (Fig.~\ref{fig:attn_vis}).
Decoder attention weights represent the weights ParaQG gives to the words in the paragraph while generating the question words. 
For example, while decoding the question word ``\textit{in}'', the system gives the highest weight to the paragraph word ``\textbf{1909}''. Similarly, the question word ``\textit{disease}'' is generated by attending over the word ``\textit{disease}'' in the paragraph.

\begin{figure}
    \centering
    \includegraphics[width=\linewidth]{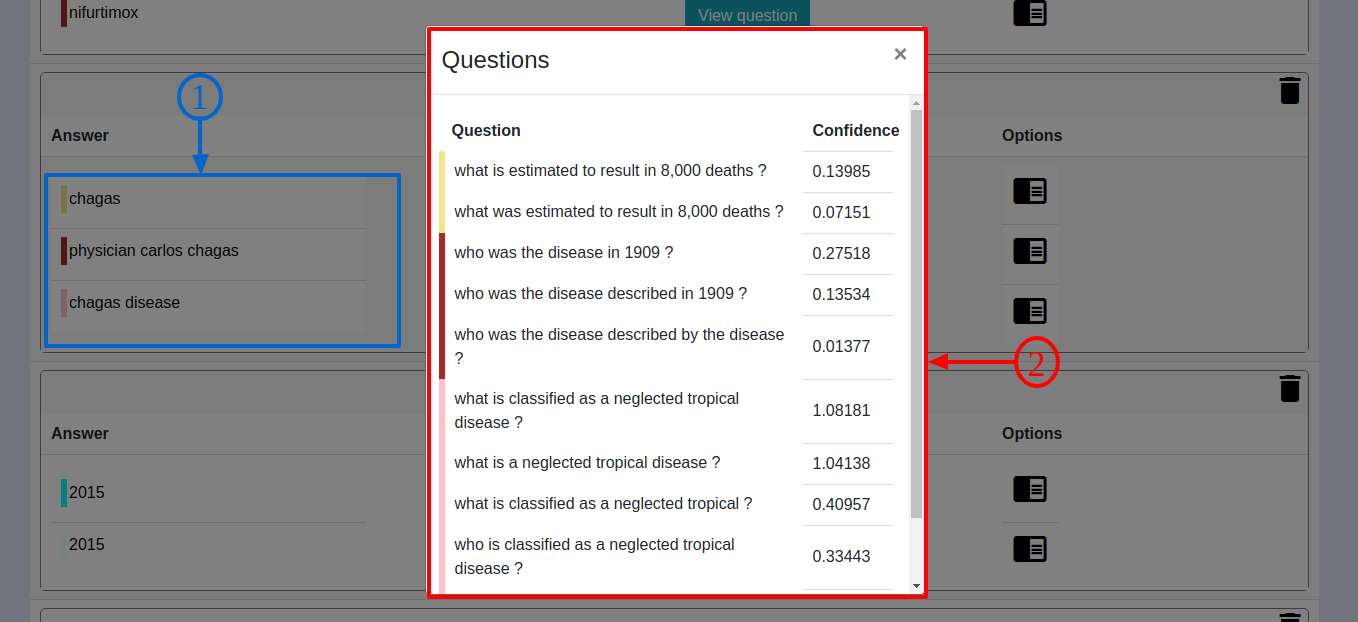}
    \caption{Filtering questions based on confidence score.}
    \label{fig:filter_ques}
\end{figure}

\begin{figure}
    \centering
    \includegraphics[width=\linewidth]{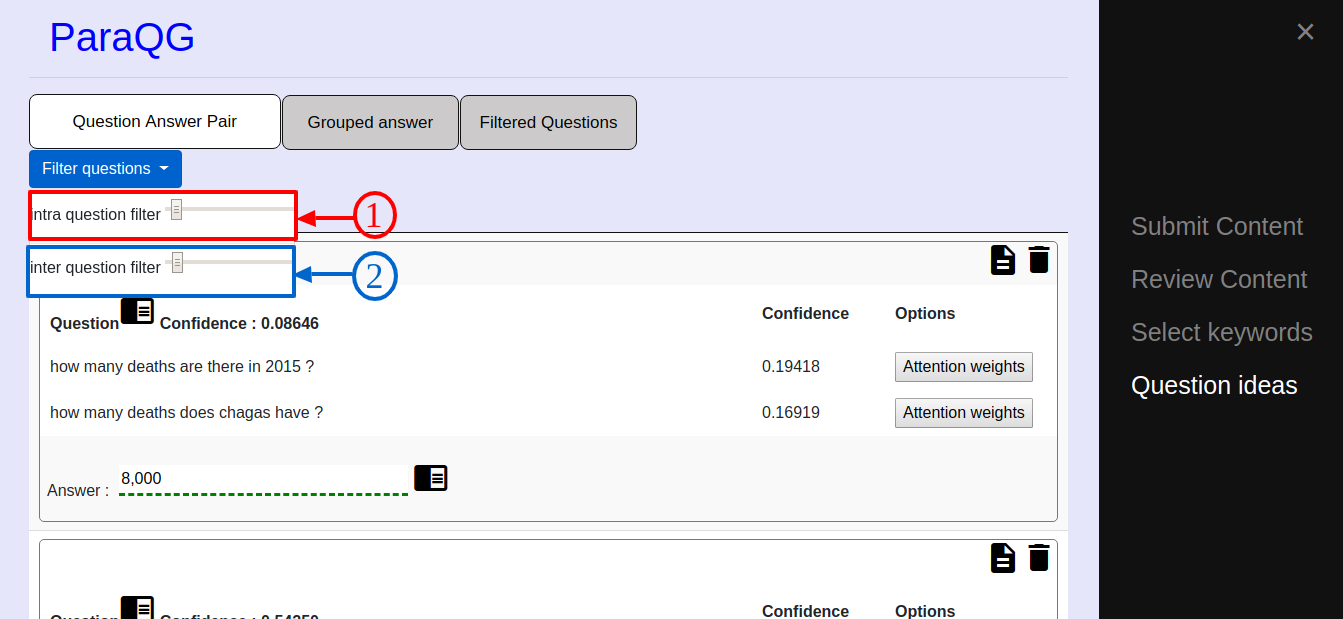}
    \caption{Clustering question based on different facets of the answer.}
    \label{fig:clustering}
\end{figure}

\paragraph{Filtering and grouping questions:} User can filter generated questions based on confidence scores using inter-question filter (Label 2 in figure \ref{fig:filter_ques}) and intra-question filter (Label 1 in Fig.~\ref{fig:filter_ques} ). The intra-question filter provides the user with a knob to filter questions based on the confidence score. The inter-question filter provides the user with  a knob to filter low quality question-answer pairs generated from the paragraph in its entirety. We filter out unanswerable questions using our BERT-based model (explained in Section~\ref{sec:filtering}). We also group answers (and thus their associated questions) based on the stemmed form of the answer. Label 1 in Fig.~\ref{fig:clustering} depicts one such answer group, whereas the generated question set is depicted by Label 2 in Fig.~\ref{fig:clustering}.

\paragraph{Editing questions with history of edits and Download generated questions and answers:}
If users are not satisfied with the generated question answer pairs he/she may edit it. The system stores all version of questions and answers. Users can download the final generated set of questions and answers in JSON or text format at the end.
\section{Implementation Details}
\label{sec:impl}
ParaQG\footnote{The source code is availble for download at \url{https://github.com/sivaanandhmuneeswaran/qg-ui}} comprises the frontend user interface, the backend question generator and a BERT-based question filtering module. The question generator model is implemented using the PyTorch\footnote{\url{https://pytorch.org/}} framework. We trained the question generator model on the SQuAD 1.0~\cite{rajpurkar2016squad} dataset. We use pre-trained GloVe word vectors of 300 dimension and fix them during training. We employ a 2-layer Bi-LSTM encoder and a single-layer Bi-LSTM decoder of hidden size 600. For optimization we use SGD with annealing. We set the initial learning rate to 0.1. We train the model for 20 epochs with batch size 64. Dropout of 0.3 was applied between vertical Bi-LSTM stacks. Our question generator module provide a REST API to which we can send requests and receive responses from in JSON format. 
The embedded Javascript is used as the template rendering engine to render the front-end of the web application along with Bootstrap 4 for responsiveness. Express is the Web application framework used for server-side on top of Node.js.
For the BERT-based filtering module, we finetune the $BERT_{base}$ model on SQuAD 2.0 for 3 epochs, and set learning rate to 3e-5 and batch size to 12. 

\section{Conclusion}
\label{conc}
Question generation from text is a task useful in many application domains, yet manual generation is labour-intensive and expensive. We presented a novel online system, ParaQG, to automatically generate questions from paragraph based on pivotal answers. The system allows users to select a set of pivotal answers and then generates a ranked set of questions for each answer. ParaQG also filters out unanswerble question using a BERT-based model. ParaQG is available as a Web application, which also incorporates a novel heat map-based visualization that shows attention weights of the decoder.

\bibliography{emnlp-ijcnlp-2019}

\begin{thebibliography}{21}
\expandafter\ifx\csname natexlab\endcsname\relax\def\natexlab#1{#1}\fi

\bibitem[{Bahdanau et~al.(2014)Bahdanau, Cho, and Bengio}]{bahdanau2014neural}
Dzmitry Bahdanau, Kyunghyun Cho, and Yoshua Bengio. 2014.
\newblock Neural machine translation by jointly learning to align and
  translate.
\newblock \emph{arXiv preprint arXiv:1409.0473}.

\bibitem[{Cho et~al.(2014)Cho, Van~Merri{\"e}nboer, Bahdanau, and
  Bengio}]{cho2014properties}
Kyunghyun Cho, Bart Van~Merri{\"e}nboer, Dzmitry Bahdanau, and Yoshua Bengio.
  2014.
\newblock On the properties of neural machine translation: Encoder-decoder
  approaches.
\newblock \emph{arXiv preprint arXiv:1409.1259}.

\bibitem[{Devlin et~al.(2018)Devlin, Chang, Lee, and
  Toutanova}]{devlin2018bert}
Jacob Devlin, Ming-Wei Chang, Kenton Lee, and Kristina Toutanova. 2018.
\newblock Bert: Pre-training of deep bidirectional transformers for language
  understanding.
\newblock \emph{arXiv preprint arXiv:1810.04805}.

\bibitem[{Du et~al.(2017)Du, Shao, and Cardie}]{xinya}
Xinya Du, Junru Shao, and Claire Cardie. 2017.
\newblock Learning to ask: Neural question generation for reading
  comprehension.
\newblock In \emph{Proceedings of the 55th ACL}, pages 1342--1352. ACL.

\bibitem[{Gu et~al.(2016)Gu, Lu, Li, and Li}]{gu2016incorporating}
Jiatao Gu, Zhengdong Lu, Hang Li, and Victor~OK Li. 2016.
\newblock Incorporating copying mechanism in sequence-to-sequence learning.
\newblock In \emph{Proceedings of the 54th ACL (Volume 1: Long Papers)},
  volume~1, pages 1631--1640.

\bibitem[{Hirschman et~al.(1999)Hirschman, Light, Breck, and
  Burger}]{hirschman1999deep}
Lynette Hirschman, Marc Light, Eric Breck, and John~D Burger. 1999.
\newblock Deep read: A reading comprehension system.
\newblock In \emph{ACL}, pages 325--332. ACL.

\bibitem[{Kumar et~al.(2019{\natexlab{a}})Kumar, Hua, Ramakrishnan, Qi, Gao,
  and Li}]{Kumar2019Difficulty}
Vishwajeet Kumar, Yuncheng Hua, Ganesh Ramakrishnan, Guilin Qi, Lianli Gao, and
  Yuan-Fang Li. 2019{\natexlab{a}}.
\newblock Difficulty-controllable multi-hop question generation from knowledge
  graphs.
\newblock In \emph{ISWC}.

\bibitem[{Kumar et~al.(2019{\natexlab{b}})Kumar, Joshi, Mukherjee,
  Ramakrishnan, and Jyothi}]{Kumar2019CrossLingualTF}
Vishwajeet Kumar, N.~Joshi, Arijit Mukherjee, Ganesh Ramakrishnan, and Preethi
  Jyothi. 2019{\natexlab{b}}.
\newblock Cross-lingual training for automatic question generation.
\newblock In \emph{ACL}.

\bibitem[{Kumar et~al.(2018)Kumar, Ramakrishnan, and Li}]{kumar18}
Vishwajeet Kumar, Ganesh Ramakrishnan, and Yuan-Fang Li. 2018.
\newblock Automating reading comprehension by generating question and answer
  pairs.
\newblock In \emph{PAKDD}. Springer.

\bibitem[{Kumar et~al.(2019{\natexlab{c}})Kumar, Ramakrishnan, and
  Li}]{Kumar2019PHBC}
Vishwajeet Kumar, Ganesh Ramakrishnan, and Yuan-Fang Li. 2019{\natexlab{c}}.
\newblock Putting the horse before the cart: A generator-evaluator framework
  for question generation from text.
\newblock \emph{SIGNLL Conference on Computational Natural Language Learning,
  CoNLL 2019}.

\bibitem[{Luong et~al.(2015)Luong, Pham, and Manning}]{LuongPM15}
Minh{-}Thang Luong, Hieu Pham, and Christopher~D. Manning. 2015.
\newblock \href {http://arxiv.org/abs/1508.04025} {Effective approaches to
  attention-based neural machine translation}.
\newblock \emph{CoRR}, abs/1508.04025.

\bibitem[{Martins and Astudillo(2016)}]{martins2016softmax}
Andre Martins and Ramon Astudillo. 2016.
\newblock From softmax to sparsemax: A sparse model of attention and
  multi-label classification.
\newblock In \emph{ICML}, pages 1614--1623.

\bibitem[{Ng et~al.(2000)Ng, Teo, and Kwan}]{ng2000machine}
Hwee~Tou Ng, Leong~Hwee Teo, and Jennifer Lai~Pheng Kwan. 2000.
\newblock A machine learning approach to answering questions for reading
  comprehension tests.
\newblock In \emph{SIGDAT-EMNLP, ACL-Volume 13}, pages 124--132. ACL.

\bibitem[{Rajpurkar et~al.(2018)Rajpurkar, Jia, and Liang}]{rajpurkar2018know}
Pranav Rajpurkar, Robin Jia, and Percy Liang. 2018.
\newblock Know what you don’t know: Unanswerable questions for squad.
\newblock In \emph{ACL}, pages 784--789.

\bibitem[{Rajpurkar et~al.(2016)Rajpurkar, Zhang, Lopyrev, and
  Liang}]{rajpurkar2016squad}
Pranav Rajpurkar, Jian Zhang, Konstantin Lopyrev, and Percy Liang. 2016.
\newblock Squad: 100,000+ questions for machine comprehension of text.
\newblock \emph{arXiv preprint arXiv:1606.05250}.

\bibitem[{Serban et~al.(2016)Serban, Garc{\'\i}a-Dur{\'a}n, Gulcehre, Ahn,
  Chandar, Courville, and Bengio}]{serban2016generating}
Iulian~Vlad Serban, Alberto Garc{\'\i}a-Dur{\'a}n, Caglar Gulcehre, Sungjin
  Ahn, Sarath Chandar, Aaron Courville, and Yoshua Bengio. 2016.
\newblock Generating factoid questions with recurrent neural networks: The 30m
  factoid question-answer corpus.
\newblock \emph{arXiv preprint arXiv:1603.06807}.

\bibitem[{Song et~al.(2018)Song, Wang, Hamza, Zhang, and
  Gildea}]{song2018leveraging}
Linfeng Song, Zhiguo Wang, Wael Hamza, Yue Zhang, and Daniel Gildea. 2018.
\newblock Leveraging context information for natural question generation.
\newblock In \emph{NAACL-HLT}, volume~2, pages 569--574.

\bibitem[{Sutskever et~al.(2014)Sutskever, Vinyals, and
  Le}]{sutskever2014sequence}
Ilya Sutskever, Oriol Vinyals, and Quoc~V Le. 2014.
\newblock Sequence to sequence learning with neural networks.
\newblock In \emph{Advances in neural information processing systems}, pages
  3104--3112.

\bibitem[{Vaswani et~al.(2017)Vaswani, Shazeer, Parmar, Uszkoreit, Jones,
  Gomez, Kaiser, and Polosukhin}]{vaswani2017attention}
Ashish Vaswani, Noam Shazeer, Niki Parmar, Jakob Uszkoreit, Llion Jones,
  Aidan~N Gomez, {\L}ukasz Kaiser, and Illia Polosukhin. 2017.
\newblock Attention is all you need.
\newblock In \emph{Advances in neural information processing systems}, pages
  5998--6008.

\bibitem[{Wolfe(1976)}]{wolfe1976automatic}
John~H Wolfe. 1976.
\newblock Automatic question generation from text-an aid to independent study.
\newblock \emph{ACM SIGCSE Bulletin}, 8(1):104--112.

\bibitem[{Zhao et~al.(2018)Zhao, Ni, Ding, and Ke}]{zhao2018paragraph}
Yao Zhao, Xiaochuan Ni, Yuanyuan Ding, and Qifa Ke. 2018.
\newblock Paragraph-level neural question generation with maxout pointer and
  gated self-attention networks.
\newblock In \emph{EMNLP 2018}, pages 3901--3910.

\end{thebibliography}
\bibliographystyle{acl_natbib}

\end{document}